\documentclass{article}

\usepackage[final]{neurips_2020}
\usepackage{natbib}
\usepackage[utf8]{inputenc} 
\usepackage[T1]{fontenc}    
\usepackage{hyperref}       
\usepackage{url}            
\usepackage{booktabs}       
\usepackage{amsfonts}       
\usepackage{nicefrac}       
\usepackage{microtype}      
\usepackage{amsmath}
\usepackage[pdftex]{graphicx}
\usepackage{xcolor}

\title{Unsupervised Regionalization of Particle-resolved Aerosol Mixing State Indices on the Global Scale}

\author{
 Zhonghua Zheng \\
  Department of Civil and Environmental Engineering\\
  University of Illinois at Urbana-Champaign\\
  Urbana, IL 61801 \\
  \texttt{zzheng25@illinois.edu} \\
  \And
 Joseph Ching \\
  Meteorological Research Institute\\
  Japan Meteorological Agency\\
  1-1 Nagamine, Tsukuba, Ibaraki, 305-0052, Japan.\\
  \texttt{jching@mri-jma.go.jp} \\
  \And
 Jeffrey H. Curtis \\
  Department of Mechanical Science and Engineering\\
  University of Illinois at Urbana-Champaign\\
  Urbana, IL 61801 \\
  \texttt{jcurtis2@illinois.edu} \\
  \And
 Yu Yao \\
  Department of Atmospheric Sciences\\
  University of Illinois at Urbana-Champaign\\
  Urbana, IL 61801 \\
  \texttt{yuyao3@illinois.edu} \\
  \And
 Peng Xu \\
  School of Environmental Science and Engineering\\
  Southern University of Science and Technology\\
  Shenzhen 518055, Guangdong Province, China\\
  \texttt{xup@sustech.edu.cn} \\
  \And
 Matthew West \\
  Department of Mechanical Science and Engineering\\
  University of Illinois at Urbana-Champaign\\
  Urbana, IL 61801 \\
  \texttt{mwest@illinois.edu} \\ 
  \And
 Nicole Riemer \\
  Department of Atmospheric Sciences\\
  University of Illinois at Urbana-Champaign\\
  Urbana, IL 61801 \\
  \texttt{nriemer@illinois.edu} \\ 
}

\begin{document}

\maketitle

\clearpage
\begin{abstract}
  The aerosol mixing state significantly affects the climate and
  health impacts of atmospheric aerosol particles. Simplified aerosol
  mixing state assumptions, common in Earth System models, can
  introduce errors in the prediction of these aerosol impacts. The
  aerosol mixing state index, a metric to quantify aerosol mixing
  state, is a convenient measure for quantifying these errors. Global
  estimates of aerosol mixing state indices have recently become
  available via supervised learning models, but require
  regionalization to ease spatiotemporal analysis. Here we developed a
  simple but effective unsupervised learning approach to regionalize
  predictions of global aerosol mixing state indices. We used the
  monthly average of aerosol mixing state indices global distribution
  as the input data. Grid cells were then clustered into regions by
  the $k$-means algorithm without explicit spatial
  information as input. This approach resulted in eleven regions over
  the globe with specific spatial aggregation patterns. Each region
  exhibited a unique distribution of mixing state indices and aerosol
  compositions, showing the effectiveness of the unsupervised
  regionalization approach. This study defines ``aerosol mixing state
  zones'' that could be useful for atmospheric science research.

\end{abstract}

\section{Introduction and Motivation}
The concept of aerosol mixing state describes how different aerosol
chemical species are distributed among the particles in a population
\citep{RIEMERETAL19}. A completely ``externally mixed'' population
contains only one species per particle, while a completely
``internally mixed'' population contains particles with the same
composition as the bulk population. Many different intermediate mixing
states are possible between those two extremes, and these commonly
occur in the ambient atmosphere. Aerosol mixing state influences the
particles' properties such as hygroscopicity
\citep{FIERCEETAL17,HOLMGRENETAL14}, optical properties
\citep{FIERCEETAL16,JACOBSON01}, cloud condensation nuclei activity
\citep{CHINGETAL12,WANGETAL10}, ice nucleation potential
\citep{KNOPFETAL18}, the aerosols' lifetime \citep{KOCHETAL09} in the
atmosphere, and their deposition in the human respiratory system
\citep{CHINGKAJINO18}. However, Earth System models (ESMs) or regional
climate models usually hold simplified mixing state assumptions. These
can influence how accurately physicochemical properties of aerosols
are predicted \citep{CHINGKAJINO18, FIERCEETAL17}, and thereby limit
the accuracy of estimating aerosol impacts on climate and health.

\citet{RIEMERWEST13} proposed an entropy-based diversity metric to
quantify aerosol mixing state. This metric, the mixing state index
($\chi$), has a range from 0\% (a fully externally mixed population)
to 100\% (a fully internally mixed population). The index $\chi$ 
has been applied to both modeling \citep{ZHUETAL16} and experimental
\citep{HEALYETAL14,OBRIENETAL15,FRAUNDETAL17,BONDYETAL18,YEETAL18}
work. For example, $\chi$ has been used for
quantifying errors in CCN concentration \citep{CHINGETAL17} prediction,
as well as the prediction of soot particles depositing in the human
respiratory system \citep{CHINGKAJINO18}.

Knowing the global distribution of aerosol mixing state index is
desirable, as it can be used for error quantification at the large
scale. But deriving this distribution directly by using a benchmarking
particle-resolved model \citep{RIEMERETAL09} at the global scale would
be computationally prohibitive over the time scales of interest. To
overcome this limitation, supervised-learning emulators were developed
for predicting the spatial distribution aerosol mixing state
across the globe \citep{HUGHESETAL18,ZHENGETAL20}. However, the
spatial delimitation based on aerosol mixing state indices remains
unclear, which hinders further spatiotemporal analysis.

The research questions for this paper are: (1) Is it possible to
regionalize the global mixing state indices, i.e., define aerosol
mixing state zones, similar to distinct climate zones
\citep{FOVELLFOVELL93,MITCHELL76}? (2) What are spatiotemporal
patterns of the regionalized mixing state indices? This paper
describes an effort to regionalize the global mixing state indices
using a simple but effective unsupervised learning approach.

\section{Methods}
\subsection{Mixing state indices definition}

The mixing state index $\chi$ \citep{RIEMERWEST13} is given by the affine
ratio of the average particle species diversity ($D_{\alpha}$) and bulk
population species diversity ($D_{\gamma}$) as
${\chi} = \frac{D_{\alpha}-1}{D_{\gamma}-1}$. 
The diversities $D_{\alpha}$ and $D_{\gamma}$ are calculated based on
per-particle species mass fractions and mass fraction of particles as
described in detail in \citet{RIEMERWEST13}.

This study focused on submicron aerosols because they are the dominant
category of particles for radiation interactions and the provision of
cloud condensation nuclei. We defined the mixing state indices in
three different ways,
namely based on chemical species abundance ($\chi_\textrm{a}$), based
on the mixing of absorbing and non-absorbing species
($\chi_\textrm{o}$), and based on the mixing of hygroscopic and
non-hygroscopic species ($\chi_\textrm{h}$). The index
$\chi_\textrm{a}$ was defined based on all the model aerosol
species. A lower value for $\chi_\textrm{a}$ indicates that the
species tend to be present in separate particles. For
$\chi_\textrm{o}$, we considered two surrogate species, black carbon
(absorbing) and all other aerosol species grouped together (assumed to
be non-absorbing). Thus, a lower value for $\chi_\textrm{o}$ indicates
the absorbing species black carbon and the sum of all other
(non-absorbing) species are more externally mixed. Similarly,
$\chi_\textrm{h}$ was also calculated from two surrogate species. We
combined black carbon, dust, and primary organic matter as one
surrogate species, given their comparatively lower hygroscopicities,
and salt, secondary organics, and sulfate as the other surrogate
species. Correspondingly, a lower value for $\chi_\textrm{h}$
represents the case where hygroscopic and non-hygroscopic species
tended to be present in separate particles. Note that mixing state
indices defined for different purposes capture different aspects of
the overall mixing state, and they are uncorrelated
\citep{ZHENGETAL20}.

\subsection{Data}
We used the supervised-learning surrogate models developed by
\citet{ZHENGETAL20} to predict the mixing state indices from ESM
(Earth System Model) output. The surrogate models were trained on an
ensemble of particle-resolved model simulations generated by the
particle-resolved model PartMC-MOSAIC
\citep{RIEMERETAL09,ZAVERIETAL08}, with the training labels for $\chi$
calculated directly from the particle-resolved data. The strategy to
generate the training and testing data was to vary 45 input parameters
for the PartMC-MOSAIC model scenarios, including primary emissions of
different aerosol types, primary emissions of gas phase species, and
meteorological parameters. A Latin Hypercube sampling approach was
employed to sample the parameter space efficiently. The surrogate
models were trained by using XGBoost (eXtreme Gradient Boosting,
\citet{CHENGUESTRIN16}), and can be expressed as:
\begin{equation}
\chi_S = f_S(A,G,E).
\end{equation}
Where $\chi_S$ is the mixing state index ($\chi_\textrm{a}$,
$\chi_\textrm{o}$, or $\chi_\textrm{h}$) and $f_S$ denotes
the surrogate function for the corresponding mixing state index.
The variants $A$ (aerosol), $G$ (gas), and $E$
(environment) represent the features (variables) from the PartMC
simulations that are used for predicting the mixing state index. These variables are
also available from the ESM.  A detailed definition of the features for the
surrogate models is given in \citet{ZHENGETAL20}.

We used the simulations from the Community Earth System Model Version
2 \citep[CESM2 version 2.1.0;][]{DANABASOGLUETAL20} with MAM4
\citep{LIUETAL16} to provide variables at the global
scale. Specifically, we used the component set ``FHIST'' for the
global simulation configuration. This component set represents a
typical historical simulation using an active atmosphere and land with
prescribed sea surface temperatures and sea-ice extent, and a 1~degree
finite volume dycore with the forcing data available from 1979 to
2015. We ran the model for the year 2011 with 6~years (2005-2010)
spinup. The simulation was conducted at a resolution of 0.9$^\circ$
latitude by 1.25$^\circ$ longitude along with emission inventories
from CMIP6 emissions \citep{EMMONSETAL20}. We stored the instantaneous
outputs every three hours in the simulation, which yielded
2920~timestamps for each surface-layer grid cell for the entire year
of simulation time.

With the surrogate models and the ESM simulation, we predicted the
mixing state indices for each grid cell at each timestamp (i.e. every
3~hours). For each grid cell, the mixing state indices were averaged by
month.  Therefore, each grid cell has a vector of stacked mixing state
indices with a length of 36 (12 months $\times$ 3 mixing state types),
represented as $\boldsymbol{x} =
(\chi_{\textrm{a},1},\dots,\chi_{\textrm{a},12},\chi_{\textrm{o},1},\dots,\chi_{\textrm{o},12},\chi_{\textrm{h},1},\dots,\chi_{\textrm{h},12})$.

\subsection{Regionalization strategy}
Here we interchangeably use the terms ``region'' and ``cluster'' to
refer to a group of grid cells. We used the $k$-means unsupervised
clustering algorithm to partition $n$ grid cells into $k$ clusters,
minimizing the within-cluster variances. The $k$-means method has
been applied to environmental sciences for ecoregion delineation
\citep{KUMARETAL11}, environmental risk zoning of the chemical
industrial area \citep{SHIZENG14}, and clustering haze trajectory of
peatland fires \citep{KHAIRATETAL17}, among other applications.

In our case, given $n=55\,296$ (192 latitude $\times$ 288 longitude) grid cells
$(\boldsymbol{x}_1, \boldsymbol{x}_2, \dots, \boldsymbol{x}_{55296})$, where
each grid cell contains a 36-dimensional vector (three mixing state indices over
12 months), $k$-means clustering aims to partition these grid cells into $k$
($k \le n$) sets $\boldsymbol{S} = \{S_1,S_2,\dots,S_k\}$ so as to
minimize the within-cluster variance. The objective is to find:
\begin{equation}
\mathop{\arg\min}_{\boldsymbol{S}} \ \ \ \sum_{i=1}^{k}\sum_{\boldsymbol{x}\in S_i}=\Vert{\boldsymbol{x}}-\boldsymbol{\mu}_i\Vert^2
\end{equation}
where $\boldsymbol{\mu}_i$ is the mean of the points in $S_i$.

The $k$-means clustering method requires the a priori specification of
the number of clusters ($k$). As the value of $k$ increases, the
variance $V_k$ asymptotically approaches 0 until $k$ equals $n$. Here
we used the relative difference in variance between $k-1$ and $k$ to
determine the number of clusters. Starting from $k=2$, we computed the
criteria of relative difference $V_{k-1} - V_{k} \le 0.05V_{k-1}$ to
determine the cluster number $k$. In this study, since
$V_{10} - V_{11} \le 0.05V_{10}$, we selected $k=11$
clusters.

\section{Results}

\begin{figure}
  \centering
  \includegraphics[width=\textwidth,trim=50 0 0 0]{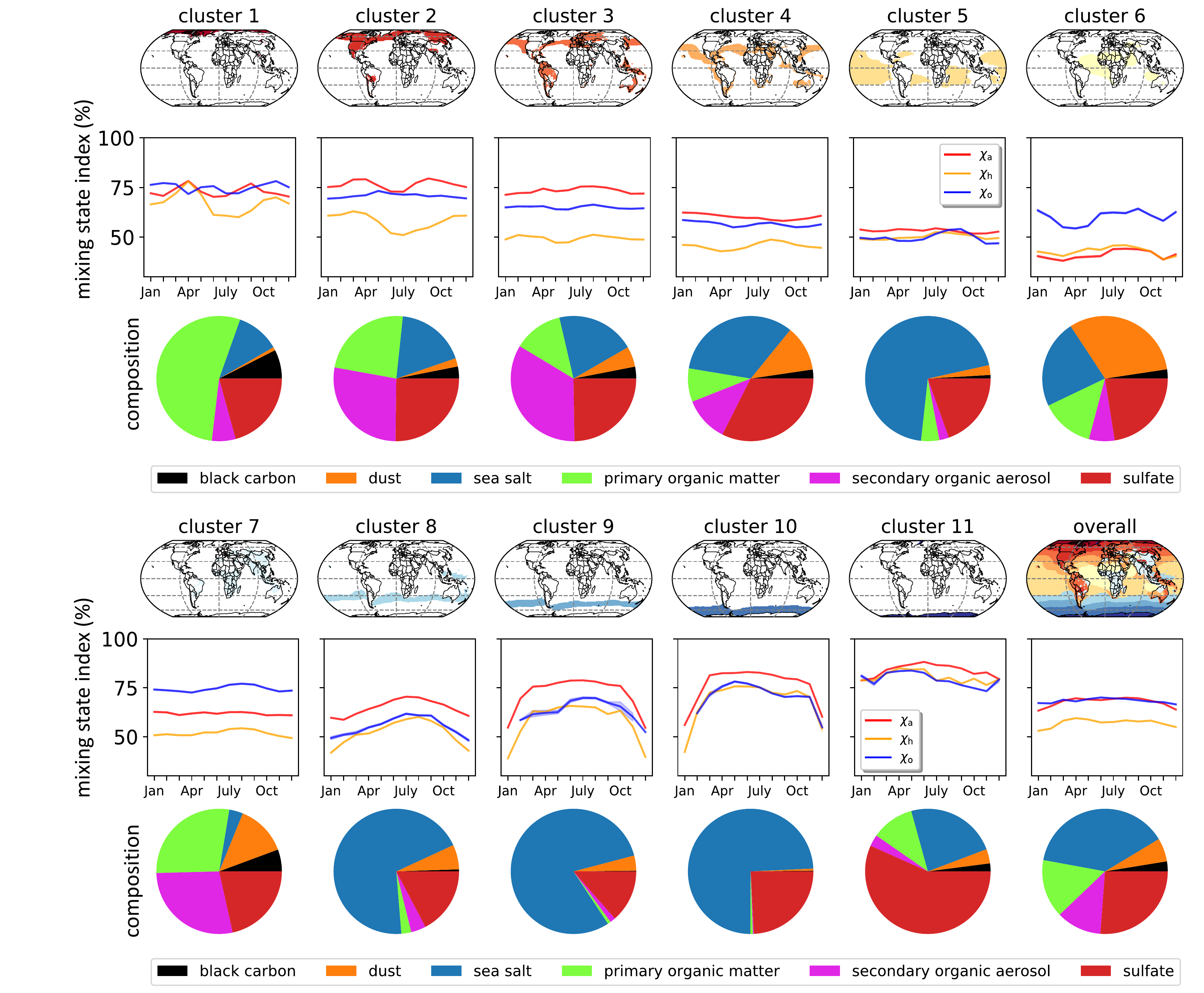}
  \caption{Aerosol mixing state zones based on unsupervised
    regionalization. Also shown are monthly average mixing state
    indices and annual averages of aerosol species mass fractions
    (bulk composition).}\label{fig:map_clusters}
\end{figure}

The regional clusters of mixing state are shown in
Figure~\ref{fig:map_clusters}. Unlike with supervised learning, the
performance of unsupervised learning are challenging to evaluate since
there is no ground truth.  However, the clustering process of this
study was merely based the information of mixing state indices,
without the direct guidance such as aerosol species distribution or
other explicit spatial information. Therefore a success indicator of
the spatiotemporal clustering is whether spatially-contiguous regions
emerge from this procedure. The clusters in
Figure~\ref{fig:map_clusters} exhibit notable regional patterns among
the eleven clusters (mixing state zones), suggesting that this
approach has indeed identified meaningful clusters. Each cluster has a
unique spatiotemporal distribution of three mixing state indices, and
captures the spatiotemporal variations that could not be detected by
the global overall averages. For example, ocean regions (e.g.,
clusters 8, 9, 10 in the Southern Hemisphere) tend to display a more
pronounced seasonal cycle than land regions (e.g., cluster 7) and
overall averages.

Clusters~1 (the Arctic) and~11 (the Antarctic) were characterized by
high mixing state indices (with monthly means of at least 67\%)
throughout the year. The high values of all three mixing state indices
means that the different aerosol species are rather internally mixed
in these regions. Clusters~5 (tropical oceans), 8 (the South China
Sea, Indian Ocean and the Pacific Ocean in the northern hemisphere
near the equator, and the northern part of the Southern Ocean) and~9
(the southern part of the Southern Ocean) were ocean areas. Similar
bulk aerosol composition was predicted in these regions, however the
annual cycle of mixing state varies more for clusters~8 and~9 than
for~5. Clusters~2, 3, and~4 covered oceans at mid and high latitudes
in the Northern Hemisphere, and lands including Northern and Southern
America, Europe, Middle East, Southeast Asia, as well as
Australia. Clusters~6 (the region
across the Atlantic from Western and Northern Africa to the northeast
coast of South Africa, as well as part of the Indian Ocean) and~7
(Middle, Eastern, and Southern Africa, as well as Asia) were
characterized by their higher values in $\chi_\textrm{o}$ compared to
$\chi_\textrm{a}$ and $\chi_\textrm{h}$, meaning that black carbon and
non-absorbing species are comparatively internally mixed.

\section{Conclusions}
In this paper, we developed a simple but effective unsupervised
learning approach to regionalize global aerosol mixing state
indices. We used the monthly averages of the spatially-varying aerosol
mixing state indices as input data. Each grid cell was then assigned
to a region using $k$-means clustering without explicit spatial
information as input. This approach resulted in eleven distinct
regions over the world with specific spatial aggregation
patterns. Each region exhibited a unique distribution of mixing state
indices, suggesting that the unsupervised regionalization approach had
identified meaningful regions. To the best of our knowledge, this is
the first study to define aerosol mixing state zones, which we
anticipate will be helpful for future studies of the global aerosol
burden.

\begin{ack}
We would like to acknowledge high-performance computing support from
Cheyenne (\url{doi:10.5065/D6RX99HX}) provided by NCAR's Computational and
Information Systems Laboratory, sponsored by the National Science
Foundation. The CESM project is supported primarily by the National
Science Foundation.
This research used resources of the Oak Ridge Leadership Computing
Facility, which is a DOE Office of Science User Facility supported
under Contract DE-AC05-00OR22725. This research was supported in part
by an appointment to the Oak Ridge National Laboratory ASTRO Program,
sponsored by the U.S. Department of Energy and administered by the Oak
Ridge Institute for Science and Education. We also acknowledge funding
from DOE grant DE-SC0019192 and NSF grant AGS-1254428.
This research is part of the Blue Waters sustained-petascale computing
project, which is supported by the National Science Foundation (awards
OCI-0725070 and ACI-1238993) the State of Illinois, and as of
December, 2019, the National Geospatial-Intelligence Agency. Blue
Waters is a joint effort of the University of Illinois at
Urbana-Champaign and its National Center for Supercomputing
Applications. Louisa Emmons is thanked for thoughful comments on the
CESM2 simulations and the manuscript. We thank AWS for providing AWS
Cloud Credits for Research.
\end{ack}

\clearpage
\bibliography{mixing_state_clustering}  

\begin{thebibliography}{31}
\providecommand{\natexlab}[1]{#1}
\providecommand{\url}[1]{\texttt{#1}}
\expandafter\ifx\csname urlstyle\endcsname\relax
  \providecommand{\doi}[1]{doi: #1}\else
  \providecommand{\doi}{doi: \begingroup \urlstyle{rm}\Url}\fi

\bibitem[Bondy et~al.(2018)Bondy, Bonanno, Moffet, Wang, Laskin, and
  Ault]{BONDYETAL18}
Amy~L. Bondy, Daniel Bonanno, Ryan~C. Moffet, Bingbing Wang, Alexander Laskin,
  and Andrew~P. Ault.
\newblock The diverse chemical mixing state of aerosol particles in the
  southeastern {{United States}}.
\newblock \emph{Atmos. Chem. Phys.}, 18\penalty0 (16):\penalty0 12595--12612,
  2018.
\newblock ISSN 1680-7324.
\newblock \doi{10.5194/acp-18-12595-2018}.

\bibitem[Chen and Guestrin(2016)]{CHENGUESTRIN16}
Tianqi Chen and Carlos Guestrin.
\newblock {{XGBoost}}: {{A Scalable Tree Boosting System}}.
\newblock In \emph{Proceedings of the 22nd {{ACM SIGKDD International
  Conference}} on {{Knowledge Discovery}} and {{Data Mining}}}, pages 785--794,
  {San Francisco, California, USA}, 2016. {ACM Press}.
\newblock ISBN 978-1-4503-4232-2.
\newblock \doi{10.1145/2939672.2939785}.

\bibitem[Ching et~al.(2012)Ching, Riemer, and West]{CHINGETAL12}
J.~Ching, N.~Riemer, and M.~West.
\newblock Impacts of black carbon mixing state on black carbon nucleation
  scavenging: {{Insights}} from a particle-resolved model.
\newblock \emph{J. Geophys. Res. Atmos.}, 117\penalty0 (D23):\penalty0
  n/a--n/a, 2012.
\newblock ISSN 01480227.
\newblock \doi{10.1029/2012JD018269}.

\bibitem[Ching and Kajino(2018)]{CHINGKAJINO18}
Joseph Ching and Mizuo Kajino.
\newblock Aerosol mixing state matters for particles deposition in human
  respiratory system.
\newblock \emph{Sci Rep}, 8\penalty0 (1):\penalty0 8864, 2018.
\newblock ISSN 2045-2322.
\newblock \doi{10.1038/s41598-018-27156-z}.

\bibitem[Ching et~al.(2017)Ching, Fast, West, and Riemer]{CHINGETAL17}
Joseph Ching, Jerome Fast, Matthew West, and Nicole Riemer.
\newblock Metrics to quantify the importance of mixing state for {{CCN}}
  activity.
\newblock \emph{Atmos. Chem. Phys.}, 17\penalty0 (12):\penalty0 7445--7458,
  2017.
\newblock ISSN 1680-7324.
\newblock \doi{10.5194/acp-17-7445-2017}.

\bibitem[Danabasoglu et~al.(2020)Danabasoglu, Lamarque, Bacmeister, Bailey,
  DuVivier, Edwards, Emmons, Fasullo, Garcia, Gettelman, Hannay, Holland,
  Large, Lauritzen, Lawrence, Lenaerts, Lindsay, Lipscomb, Mills, Neale,
  Oleson, Otto-Bliesner, Phillips, Sacks, Tilmes, Kampenhout, Vertenstein,
  Bertini, Dennis, Deser, Fischer, Fox-Kemper, Kay, Kinnison, Kushner, Larson,
  Long, Mickelson, Moore, Nienhouse, Polvani, Rasch, and
  Strand]{DANABASOGLUETAL20}
G.~Danabasoglu, J.-F. Lamarque, J.~Bacmeister, D.~A. Bailey, A.~K. DuVivier,
  J.~Edwards, L.~K. Emmons, J.~Fasullo, R.~Garcia, A.~Gettelman, C.~Hannay,
  M.~M. Holland, W.~G. Large, P.~H. Lauritzen, D.~M. Lawrence, J.~T.~M.
  Lenaerts, K.~Lindsay, W.~H. Lipscomb, M.~J. Mills, R.~Neale, K.~W. Oleson,
  B.~Otto-Bliesner, A.~S. Phillips, W.~Sacks, S.~Tilmes, L.~Kampenhout,
  M.~Vertenstein, A.~Bertini, J.~Dennis, C.~Deser, C.~Fischer, B.~Fox-Kemper,
  J.~E. Kay, D.~Kinnison, P.~J. Kushner, V.~E. Larson, M.~C. Long,
  S.~Mickelson, J.~K. Moore, E.~Nienhouse, L.~Polvani, P.~J. Rasch, and W.~G.
  Strand.
\newblock The {{Community Earth System Model Version}} 2 ({{CESM2}}).
\newblock \emph{J. Adv. Model. Earth Syst.}, 12\penalty0 (2), 2020.
\newblock ISSN 1942-2466.
\newblock \doi{10.1029/2019MS001916}.

\bibitem[Emmons et~al.(2020)Emmons, Schwantes, Orlando, Tyndall, Kinnison,
  Lamarque, Marsh, Mills, Tilmes, Bardeen, Buchholz, Conley, Gettelman, Garcia,
  Simpson, Blake, Meinardi, and P{\'e}tron]{EMMONSETAL20}
Louisa~K. Emmons, Rebecca~H. Schwantes, John~J. Orlando, Geoff Tyndall, Douglas
  Kinnison, Jean-Fran{\c c}ois Lamarque, Daniel Marsh, Michael~J. Mills, Simone
  Tilmes, Charles Bardeen, Rebecca~R. Buchholz, Andrew Conley, Andrew
  Gettelman, Rolando Garcia, Isobel Simpson, Donald~R. Blake, Simone Meinardi,
  and Gabrielle P{\'e}tron.
\newblock The {{Chemistry Mechanism}} in the {{Community Earth System Model
  Version}} 2 ({{CESM2}}).
\newblock \emph{J. Adv. Model. Earth Syst.}, 12\penalty0 (4), 2020.
\newblock ISSN 1942-2466.
\newblock \doi{10.1029/2019MS001882}.

\bibitem[Fierce et~al.(2016)Fierce, Bond, Bauer, Mena, and
  Riemer]{FIERCEETAL16}
Laura Fierce, Tami~C. Bond, Susanne~E. Bauer, Francisco Mena, and Nicole
  Riemer.
\newblock Black carbon absorption at the global scale is affected by
  particle-scale diversity in composition.
\newblock \emph{Nat Commun}, 7\penalty0 (1):\penalty0 12361, 2016.
\newblock ISSN 2041-1723.
\newblock \doi{10.1038/ncomms12361}.

\bibitem[Fierce et~al.(2017)Fierce, Riemer, and Bond]{FIERCEETAL17}
Laura Fierce, Nicole Riemer, and Tami~C. Bond.
\newblock Toward {{Reduced Representation}} of {{Mixing State}} for
  {{Simulating Aerosol Effects}} on {{Climate}}.
\newblock \emph{Bull. Amer. Meteor. Soc.}, 98\penalty0 (5):\penalty0 971--980,
  2017.
\newblock ISSN 0003-0007, 1520-0477.
\newblock \doi{10.1175/BAMS-D-16-0028.1}.

\bibitem[Fovell and Fovell(1993)]{FOVELLFOVELL93}
Robert~G. Fovell and Mei-Ying~C. Fovell.
\newblock Climate zones of the conterminous united states defined using cluster
  analysis.
\newblock \emph{J. Clim.}, 6\penalty0 (11):\penalty0 2103--2135, 1993.
\newblock ISSN 0894-8755.
\newblock \doi{10.1175/1520-0442(1993)006<2103:CZOTCU>2.0.CO;2}.

\bibitem[Fraund et~al.(2017)Fraund, Pham, Bonanno, Harder, Wang, Brito, {de
  S{\'a}}, Carbone, China, Artaxo, Martin, P{\"o}hlker, Andreae, Laskin,
  Gilles, and Moffet]{FRAUNDETAL17}
Matthew Fraund, Don Pham, Daniel Bonanno, Tristan Harder, Bingbing Wang, Joel
  Brito, Suzane {de S{\'a}}, Samara Carbone, Swarup China, Paulo Artaxo, Scot
  Martin, Christopher P{\"o}hlker, Meinrat Andreae, Alexander Laskin, Mary
  Gilles, and Ryan Moffet.
\newblock Elemental {{Mixing State}} of {{Aerosol Particles Collected}} in
  {{Central Amazonia}} during {{GoAmazon2014}}/15.
\newblock \emph{Atmosphere}, 8\penalty0 (12):\penalty0 173, 2017.
\newblock ISSN 2073-4433.
\newblock \doi{10.3390/atmos8090173}.

\bibitem[Healy et~al.(2014)Healy, Riemer, Wenger, Murphy, West, Poulain,
  Wiedensohler, O'Connor, McGillicuddy, Sodeau, and Evans]{HEALYETAL14}
R.~M. Healy, N.~Riemer, J.~C. Wenger, M.~Murphy, M.~West, L.~Poulain,
  A.~Wiedensohler, I.~P. O'Connor, E.~McGillicuddy, J.~R. Sodeau, and G.~J.
  Evans.
\newblock Single particle diversity and mixing state measurements.
\newblock \emph{Atmos. Chem. Phys.}, 14\penalty0 (12):\penalty0 6289--6299,
  2014.
\newblock ISSN 1680-7324.
\newblock \doi{10.5194/acp-14-6289-2014}.

\bibitem[Holmgren et~al.(2014)Holmgren, Sellegri, Hervo, Rose, Freney, Villani,
  and Laj]{HOLMGRENETAL14}
H.~Holmgren, K.~Sellegri, M.~Hervo, C.~Rose, E.~Freney, P.~Villani, and P.~Laj.
\newblock Hygroscopic properties and mixing state of aerosol measured at the
  high-altitude site {{Puy}} de {{D{\^o}me}} (1465 m a.s.l.), {{France}}.
\newblock \emph{Atmos. Chem. Phys.}, 14\penalty0 (18):\penalty0 9537--9554,
  2014.
\newblock ISSN 1680-7324.
\newblock \doi{10.5194/acp-14-9537-2014}.

\bibitem[Hughes et~al.(2018)Hughes, Kodros, Pierce, West, and
  Riemer]{HUGHESETAL18}
Michael Hughes, John Kodros, Jeffrey Pierce, Matthew West, and Nicole Riemer.
\newblock Machine {{Learning}} to {{Predict}} the {{Global Distribution}} of
  {{Aerosol Mixing State Metrics}}.
\newblock \emph{Atmosphere}, 9\penalty0 (1):\penalty0 15, 2018.
\newblock ISSN 2073-4433.
\newblock \doi{10.3390/atmos9010015}.

\bibitem[Jacobson(2001)]{JACOBSON01}
Mark~Z. Jacobson.
\newblock Strong radiative heating due to the mixing state of black carbon in
  atmospheric aerosols.
\newblock \emph{Nature}, 409\penalty0 (6821):\penalty0 695--697, 2001.
\newblock ISSN 0028-0836, 1476-4687.
\newblock \doi{10.1038/35055518}.

\bibitem[Khairat et~al.(2017)Khairat, Sitanggang, and Nuryanto]{KHAIRATETAL17}
H.~Khairat, I.~Sukaesih Sitanggang, and D.~E. Nuryanto.
\newblock Clustering {{Haze Trajectory}} of {{Peatland Fires}} in {{Riau
  Province}} using {{K}}-{{Means Algorithm}}.
\newblock \emph{IOP Conf. Ser.: Earth Environ. Sci.}, 58:\penalty0 012059,
  2017.
\newblock ISSN 1755-1315.
\newblock \doi{10.1088/1755-1315/58/1/012059}.

\bibitem[Knopf et~al.(2018)Knopf, Alpert, and Wang]{KNOPFETAL18}
Daniel~A. Knopf, Peter~A. Alpert, and Bingbing Wang.
\newblock The {{Role}} of {{Organic Aerosol}} in {{Atmospheric Ice
  Nucleation}}: {{A Review}}.
\newblock \emph{ACS Earth Space Chem.}, 2\penalty0 (3):\penalty0 168--202,
  2018.
\newblock \doi{10.1021/acsearthspacechem.7b00120}.

\bibitem[Koch et~al.(2009)Koch, Schulz, Kinne, McNaughton, Spackman, Balkanski,
  Bauer, Berntsen, Bond, Boucher, Chin, Clarke, De~Luca, Dentener, Diehl,
  Dubovik, Easter, Fahey, Feichter, Fillmore, Freitag, Ghan, Ginoux, Gong,
  Horowitz, Iversen, Kirkev\&amp;aring;g, Klimont, Kondo, Krol, Liu, Miller,
  Montanaro, Moteki, Myhre, Penner, Perlwitz, Pitari, Reddy, Sahu, Sakamoto,
  Schuster, Schwarz, Seland, Stier, Takegawa, Takemura, Textor, {van Aardenne},
  and Zhao]{KOCHETAL09}
D.~Koch, M.~Schulz, S.~Kinne, C.~McNaughton, J.~R. Spackman, Y.~Balkanski,
  S.~Bauer, T.~Berntsen, T.~C. Bond, O.~Boucher, M.~Chin, A.~Clarke,
  N.~De~Luca, F.~Dentener, T.~Diehl, O.~Dubovik, R.~Easter, D.~W. Fahey,
  J.~Feichter, D.~Fillmore, S.~Freitag, S.~Ghan, P.~Ginoux, S.~Gong,
  L.~Horowitz, T.~Iversen, A.~Kirkev\&amp;aring;g, Z.~Klimont, Y.~Kondo,
  M.~Krol, X.~Liu, R.~Miller, V.~Montanaro, N.~Moteki, G.~Myhre, J.~E. Penner,
  J.~Perlwitz, G.~Pitari, S.~Reddy, L.~Sahu, H.~Sakamoto, G.~Schuster, J.~P.
  Schwarz, {\O}.~Seland, P.~Stier, N.~Takegawa, T.~Takemura, C.~Textor, J.~A.
  {van Aardenne}, and Y.~Zhao.
\newblock Evaluation of black carbon estimations in global aerosol models.
\newblock \emph{Atmos. Chem. Phys.}, 9\penalty0 (22):\penalty0 9001--9026,
  2009.
\newblock ISSN 1680-7324.
\newblock \doi{10.5194/acp-9-9001-2009}.

\bibitem[Kumar et~al.(2011)Kumar, Mills, Hoffman, and Hargrove]{KUMARETAL11}
Jitendra Kumar, Richard~T. Mills, Forrest~M. Hoffman, and William~W. Hargrove.
\newblock Parallel k-{{Means Clustering}} for {{Quantitative Ecoregion
  Delineation Using Large Data Sets}}.
\newblock \emph{Procedia Computer Science}, 4:\penalty0 1602--1611, 2011.
\newblock ISSN 18770509.
\newblock \doi{10.1016/j.procs.2011.04.173}.

\bibitem[Liu et~al.(2016)Liu, Ma, Wang, Tilmes, Singh, Easter, Ghan, and
  Rasch]{LIUETAL16}
X.~Liu, P.-L. Ma, H.~Wang, S.~Tilmes, B.~Singh, R.~C. Easter, S.~J. Ghan, and
  P.~J. Rasch.
\newblock Description and evaluation of a new four-mode version of the {{Modal
  Aerosol Module}} ({{MAM4}}) within version 5.3 of the {{Community Atmosphere
  Model}}.
\newblock \emph{Geosci. Model Dev.}, 9\penalty0 (2):\penalty0 505--522, 2016.
\newblock ISSN 1991-9603.
\newblock \doi{10.5194/gmd-9-505-2016}.

\bibitem[Mitchell(1976)]{MITCHELL76}
Val~L. Mitchell.
\newblock The regionalization of climate in the western united states.
\newblock \emph{J. Appl. Meteorol.}, 15\penalty0 (9):\penalty0 920--927, 1976.
\newblock ISSN 0021-8952.
\newblock \doi{10.1175/1520-0450(1976)015<0920:TROCIT>2.0.CO;2}.

\bibitem[O'Brien et~al.(2015)O'Brien, Wang, Laskin, Riemer, West, Zhang, Sun,
  Yu, Alpert, Knopf, Gilles, and Moffet]{OBRIENETAL15}
Rachel~E. O'Brien, Bingbing Wang, Alexander Laskin, Nicole Riemer, Matthew
  West, Qi~Zhang, Yele Sun, Xiao-Ying Yu, Peter Alpert, Daniel~A. Knopf,
  Mary~K. Gilles, and Ryan~C. Moffet.
\newblock Chemical imaging of ambient aerosol particles: {{Observational}}
  constraints on mixing state parameterization.
\newblock \emph{J. Geophys. Res. Atmos.}, 120\penalty0 (18):\penalty0
  9591--9605, 2015.
\newblock ISSN 2169-897X, 2169-8996.
\newblock \doi{10.1002/2015JD023480}.

\bibitem[Riemer and West(2013)]{RIEMERWEST13}
N.~Riemer and M.~West.
\newblock Quantifying aerosol mixing state with entropy and diversity measures.
\newblock \emph{Atmos. Chem. Phys.}, 13\penalty0 (22):\penalty0 11423--11439,
  2013.
\newblock ISSN 1680-7324.
\newblock \doi{10.5194/acp-13-11423-2013}.

\bibitem[Riemer et~al.(2009)Riemer, West, Zaveri, and Easter]{RIEMERETAL09}
N.~Riemer, M.~West, R.~A. Zaveri, and R.~C. Easter.
\newblock Simulating the evolution of soot mixing state with a
  particle-resolved aerosol model.
\newblock \emph{J. Geophys. Res. Atmos.}, 114\penalty0 (D9):\penalty0 D09202,
  2009.
\newblock ISSN 0148-0227.
\newblock \doi{10.1029/2008JD011073}.

\bibitem[Riemer et~al.(2019)Riemer, Ault, West, Craig, and
  Curtis]{RIEMERETAL19}
N.~Riemer, A.~P. Ault, M.~West, R.~L. Craig, and J.~H. Curtis.
\newblock Aerosol {{Mixing State}}: {{Measurements}}, {{Modeling}}, and
  {{Impacts}}.
\newblock \emph{Rev. Geophys.}, 57\penalty0 (2):\penalty0 187--249, 2019.
\newblock ISSN 8755-1209, 1944-9208.
\newblock \doi{10.1029/2018RG000615}.

\bibitem[Shi and Zeng(2014)]{SHIZENG14}
Weifang Shi and Weihua Zeng.
\newblock Application of k-means clustering to environmental risk zoning of the
  chemical industrial area.
\newblock \emph{Front. Environ. Sci. Eng.}, 8\penalty0 (1):\penalty0 117--127,
  2014.
\newblock ISSN 2095-2201, 2095-221X.
\newblock \doi{10.1007/s11783-013-0581-5}.

\bibitem[Wang et~al.(2010)Wang, Cubison, Aiken, Jimenez, and
  Collins]{WANGETAL10}
J.~Wang, M.~J. Cubison, A.~C. Aiken, J.~L. Jimenez, and D.~R. Collins.
\newblock The importance of aerosol mixing state and size-resolved composition
  on {{CCN}} concentration and the variation of the importance with atmospheric
  aging of aerosols.
\newblock \emph{Atmos. Chem. Phys.}, 10\penalty0 (15):\penalty0 7267--7283,
  2010.
\newblock ISSN 1680-7324.
\newblock \doi{10.5194/acp-10-7267-2010}.

\bibitem[Ye et~al.(2018)Ye, Gu, Li, Robinson, Lipsky, Kaltsonoudis, Lee, Apte,
  Robinson, Sullivan, Presto, and Donahue]{YEETAL18}
Qing Ye, Peishi Gu, Hugh~Z. Li, Ellis~S. Robinson, Eric Lipsky, Christos
  Kaltsonoudis, Alex~K.Y. Lee, Joshua~S. Apte, Allen~L. Robinson, Ryan~C.
  Sullivan, Albert~A. Presto, and Neil~M. Donahue.
\newblock Spatial {{Variability}} of {{Sources}} and {{Mixing State}} of
  {{Atmospheric Particles}} in a {{Metropolitan Area}}.
\newblock \emph{Environ. Sci. Technol.}, 52\penalty0 (12):\penalty0 6807--6815,
  2018.
\newblock ISSN 0013-936X.
\newblock \doi{10.1021/acs.est.8b01011}.

\bibitem[Zaveri et~al.(2008)Zaveri, Easter, Fast, and Peters]{ZAVERIETAL08}
Rahul~A. Zaveri, Richard~C. Easter, Jerome~D. Fast, and Leonard~K. Peters.
\newblock Model for {{Simulating Aerosol Interactions}} and {{Chemistry}}
  ({{MOSAIC}}).
\newblock \emph{J. Geophys. Res. Atmos.}, 113\penalty0 (D13), 2008.
\newblock ISSN 2156-2202.
\newblock \doi{10.1029/2007JD008782}.

\bibitem[Zheng et~al.(2020)Zheng, Curtis, Yao, Gasparik, Anantharaj, Zhao,
  West, and Riemer]{ZHENGETAL20}
Zhonghua Zheng, Jeffrey~H. Curtis, Yu~Yao, Jessica~T. Gasparik, Valentine~G.
  Anantharaj, Lei Zhao, Matthew West, and Nicole Riemer.
\newblock Estimating {{Submicron Aerosol Mixing State}} at the {{Global Scale}}
  with {{Machine Learning}} and {{Earth System Modeling}}.
\newblock Preprint, {EarthArXiv}, 2020.

\bibitem[Zhu et~al.(2016)Zhu, Sartelet, Zhang, and Nenes]{ZHUETAL16}
Shupeng Zhu, Karine Sartelet, Yang Zhang, and Athanasios Nenes.
\newblock Three-dimensional modeling of the mixing state of particles over
  {{Greater Paris}}: 3-{{D MIXING STATE MODELING}}.
\newblock \emph{J. Geophys. Res. Atmos.}, 121\penalty0 (10):\penalty0
  5930--5947, 2016.
\newblock ISSN 2169897X.
\newblock \doi{10.1002/2015JD024241}.

\end{thebibliography}

\end{document}